\documentclass[letterpaper, 10 pt, conference]{ieeeconf}  

\IEEEoverridecommandlockouts                              

\usepackage{amsmath} 
\usepackage{amssymb}  
\usepackage{dsfont}
\usepackage{graphicx}

\usepackage{psfrag,graphicx,epsfig}
\usepackage{epstopdf}
\usepackage{xspace}
\usepackage{subfig}
\usepackage{float}
\usepackage{placeins}
\usepackage{multirow}
\usepackage{pgf,tikz}
\usepackage{nowidow}
\usepackage{lineno}
\usepackage{xcolor}
\usepackage{lipsum}
\usepackage{dsfont}

\usepackage{siunitx}
\usepackage{color}
\usepackage{flushend}
\usepackage{lineno}
\usepackage{algorithm}
\usepackage[noend]{algorithmic}
\usepackage{booktabs}
\allowdisplaybreaks

\definecolor{myred}{rgb}{0.792, 0, 0.125}
\definecolor{myorange}{rgb}{0.902, 0.380, 0.004}
\definecolor{mypurple}{rgb}{0.369, 0.235, 0.6}
\definecolor{myblue}{rgb}{0.020, 0.443, 0.690}
\definecolor{mygray_d}{rgb}{0.188, 0.188, 0.188}
\definecolor{mygray_l}{rgb}{0.563, 0.563, 0.563}

\title{\LARGE \bf \textsc{\textbf{FDPP}}: Fine-tune Diffusion Policy with Human Preference}

\author{Yuxin Chen$^{1\dag}$, Devesh K. Jha$^{2}$, Masayoshi Tomizuka$^{1}$, Diego Romeres$^{2}$
\thanks{$^1$Mechanical Systems Control Lab, UC Berkeley, Berkeley, CA, USA {\tt\small \{yuxinc, tomizuka\}}@berkeley.edu}
\thanks{$^2$Mitsubishi Electric Research Laboratories (MERL), Cambridge, MA, USA {\tt\small \{jha, romeres\}}@merl.com}
\thanks{$^\dag$Work done during MERL internship.}}

\begin{document}

\maketitle


\begin{abstract}
Imitation learning from human demonstrations enables robots to perform complex manipulation tasks and has recently witnessed huge success. However, these techniques often struggle to adapt behavior to new preferences or changes in the environment. To address these limitations, we propose Fine-tuning Diffusion Policy with Human Preference (FDPP). FDPP learns a reward function through preference-based learning. This reward is then used to fine-tune the pre-trained policy with reinforcement learning (RL), resulting in alignment of pre-trained policy with new human preferences while still solving the original task. Our experiments across various robotic tasks and preferences demonstrate that FDPP effectively customizes policy behavior without compromising performance. Additionally, we show that incorporating Kullback–Leibler (KL) regularization during fine-tuning prevents over-fitting and helps maintain the competencies of the initial policy.
\end{abstract}
\section{Introduction}\label{sec:intro}

Imitation learning from human demonstrations is a powerful method for training robots to perform a wide range of manipulation tasks, such as grasping~\cite{wangmimicplay,ze2023gnfactor,chi2023diffusion}, dexterous manipulation~\cite{qin2022dexmv,arunachalam2023dexterous}, and legged locomotion~\cite{peng2020learning}. Recently, the rapid advancement of generative models has highlighted their remarkable ability to synthesize complex, high-dimensional distributions, offering new opportunities for enhancing policy learning~\cite{chi2023diffusion}. Among these, diffusion models, a type of generative model that gradually transform random noise into a data sample, have been applied in imitation learning for robotics. These models, referred to as \emph{diffusion policies}, have achieved state-of-the-art performance by leveraging the powerful generative modeling capabilities~\cite{chi2023diffusion,ze20243d}.

However, diffusion policies share common challenges with general imitation learning methods. For example, training a robust diffusion policy~\cite{chi2023diffusion} for a particular task typically requires 100 to 200 human-collected demonstrations, making the process both time-consuming and sample-inefficient. Additionally, these policies are task-specific, necessitating a new set of demonstrations for each task. The environmental setup during demonstration must also closely resemble the deployment environment in terms of viewpoint, object appearance, and action space~\cite{ze20243d}. Nevertheless, during real-world deployment, it is common to encounter additional constraints (e.g., avoiding undesired regions during movement) or preferences (e.g., aligning blocks rather than unstable stacking during a block stacking task) that differ from the pre-collected demonstrations. This mismatch between the policy's learned behavior and the task requirements creates a significant challenge. Therefore, effectively adapting a pre-trained diffusion policy to new environments is essential for successful real-world deployment.

\begin{figure}[t]
    \centering
    \includegraphics[width=\linewidth]{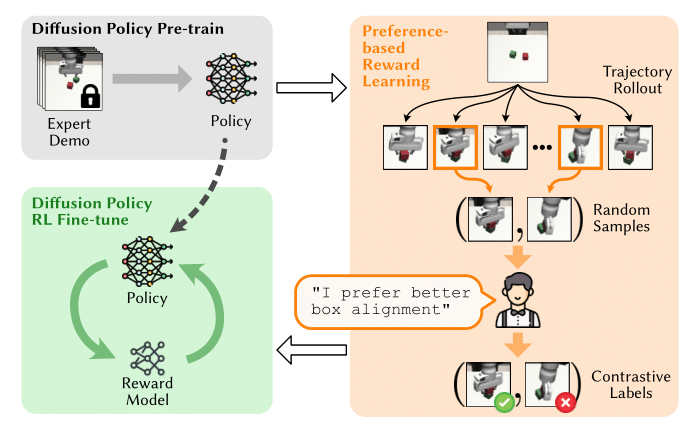}
    \caption{\textbf{Fine-tune Diffusion Policy with Human Preference.} Given a pre-trained diffusion policy, \textsc{\textbf{FDPP}} collects trajectory roll-outs and queries human feedback to label pairs of randomly sampled image observations based on human preferences or task specifications. Using these labels, a reward function is trained through preference-based reward learning, which is then used to fine-tune the diffusion policy via reinforcement learning.}
    \label{fig:main}
\end{figure}

Motivated by this challenging problem, we propose fine-tuning the diffusion policy using online reinforcement learning (RL) to better align with human preferences and task specifications. We introduce \textbf{F}ine-tuning \textbf{D}iffusion \textbf{P}olicy with Human \textbf{P}reference (\textsc{\textbf{FDPP}}), a straightforward yet effective algorithm that learns reward functions through preference-based learning with human labels. The pre-trained policy is then fine-tuned using the learned reward function via RL. An overview of these steps is illustrated in Fig.~\ref{fig:main}. We evaluate \textsc{\textbf{FDPP}} on a variety of robotic tasks with differing preferences and find that it effectively adapts the behavioral distribution of the pre-trained diffusion policy to align with human preferences, without compromising performance on the original task.

In summary, we make the following contributions:
\begin{itemize}
    \item We propose \textsc{\textbf{FDPP}}, a method that fine-tune the pre-trained diffusion policy to align with human preference.
    \item We investigate how incorporating Kullback–Leibler (KL) regularization into diffusion policy fine-tuning can effectively prevent over-fitting to the reward while preserving the original performance of the pre-trained diffusion policy on the original tasks.
\end{itemize}
We conduct empirical evaluations of \textsc{\textbf{FDPP}} across various robotic tasks and preferences, highlighting these contributions in Sec.~\ref{sec:experiments}.
\section{Related Work}\label{sec:related_works}
\subsection{Diffusion Policy}

Diffusion policy, initially introduced by Chi et al.~\cite{chi2023diffusion}, represents a significant advancement in imitation learning by leveraging generative models, specifically diffusion models~\cite{ho2020denoising,sohl2015deep,song2020score}, to replicate complex behaviors from demonstrations. This approach takes in the most recent observations, either the low-dimensional state representations or high-dimensional images, and outputs a sequence of future actions over a prediction horizon. It has achieved state-of-the-art results in various robotic tasks. Building on this, Ze et al.~\cite{ze20243d} developed the 3D diffusion policy, which incorporates 3D visual representations (such as 3D point clouds) to enhance generalizability and effectiveness compared to the original diffusion policy. However, despite their success, both diffusion policies and their variants share limitations typical of traditional imitation learning, such as the need for large amounts of task-specific demonstrations and sensitivity to environmental changes between training and deployment. Compared to this line of works, we endeavor to integrate RL fine-tuning to expand the applicability of diffusion policies in practical scenarios, where robustness and adaptability are essential.

\subsection{Preference-based Reward Learning}
In the past, demonstrations have been the preferred method for reward learning. A popular paradigm is inverse reinforcement learning (IRL)~\cite{ng2000algorithms}, where a reward function is extracted to capture why specific behaviors may be desirable from the demonstration. Recently, there is a growing trend toward using preference-based learning~\cite{christiano2017deep,sadigh2017active,biyik2018batch,wirth2017survey,brown2019extrapolating,stiennon2020learning,zhang2022time,shin2023benchmarks}. In the proposed approach, humans are asked to compare two (or more) trajectories (or states) and provide labels, allowing the model to infer a mapping from these ranked trajectories to a scalar reward. In general, human preferences and rankings of robot trajectories are easier for people to provide than kinesthetic teaching or detailed feedback~\cite{shin2023benchmarks,tian2023matters}. We leverage these approaches in our method to derive a reward function aligned with human preferences.

\subsection{RL-based Fine-tuning of Diffusion Models}
There are multiple strategies for fine-tuning diffusion models. Fan and Lee~\cite{fan2023optimizing} are the first to introduce the idea of fine-tuning pre-trained diffusion models by combining policy gradients with generative adversarial networks (GANs)~\cite{goodfellow2020generative}. In their approach, policy gradient updates are guided by reward signals from the discriminator of GAN to refine the diffusion model. They demonstrate that this fine-tuning method enables the model to generate realistic samples with fewer diffusion steps, particularly when using denoising diffusion probabilistic models (DDPMs)~\cite{ho2020denoising} sampling in simpler domains. More recently, Black et al.~\cite{black2023training} and Fan et al.~\cite{fan2024reinforcement} have proposed fine-tuning text-to-image diffusion models using RL. Both studies treat the fine-tuning process as a multi-step decision-making problem and show that RL-based fine-tuning can surpass supervised fine-tuning methods that rely on reward-weighted loss~\cite{lee2023aligning}. Fan et al. further provide a detailed analysis of KL-regularization for both supervised and RL-based fine-tuning, supported by theoretical justifications. In contrast to previous work, which primarily focuses on text-to-image diffusion models, our approach extends the application to diffusion policies in the context of robotic tasks.

Ren et al. introduce DPPO in a concurrent study~\cite{ren2024diffusion} focusing on fine-tuning a diffusion policy using the policy gradient method to enhance training stability and policy robustness. The reward for fine-tuning remains tied to the original task objective. In contrast, our method fine-tunes the policy using a new reward derived from human preferences, which may differ from the original task objective. This necessitates the use of KL regularization to prevent over-fitting and preserve the original performance of diffusion policy. Furthermore, we demonstrate how to use a preference-based reward model with contrastive labels to obtain this reward function from human preference, effectively aligning the policy with user expectations.
\section{Preliminaries}\label{sec:preliminaries}
In this section, we present a concise overview of the RL problem formulation (Sec.~\ref{sec:MDP_and_RL}) and the diffusion policy framework (Sec.~\ref{sec:dm_and_dp}).

\subsection{Markov Decision Process and Reinforcement Learning}\label{sec:MDP_and_RL}
A Markov Decision Process (MDP) defined by the tuple $\mathcal{M}=\left(\mathcal{S},\mathcal{A},r,p,\rho_0\right)$, where $\mathcal{S}\in\mathbb{R}^S$ represents the state space, $\mathcal{A}\in\mathbb{R}^A$ is the action space, $r:\mathcal{S}\times\mathcal{A}\mapsto\mathbb{R}$ is the reward function, $p:\mathcal{S}\times\mathcal{A}\times\mathcal{S}\mapsto[0,\infty)$ defines the probability density of the next state $\mathbf{s}_{t+1}\in\mathcal{S}$, given the current state $\mathbf{s}_{t}\in\mathcal{S}$ and action $\mathbf{a}_{t}\in\mathcal{A}$. The initial state distribution is denoted by $\rho_0$. At each time step $t$, the robot agent observes the state $\mathbf{s}_t$, selects an action $\mathbf{a}_t$, receives a reward $r\left(\mathbf{s}_t,\mathbf{a}_t\right)$, and transitions to the next state $\mathbf{s}_{t+1}$ following the transition probability $p\left(\mathbf{s}_{t+1}|\mathbf{s}_t,\mathbf{a}_t\right)$. 

With a given policy $\pi_\theta\left(\mathbf{a}|\mathbf{s}\right)$, parameterized by $\theta$, and the initial state $\mathbf{s}_0\sim\rho_0$, the robot agent generates a trajectory, which is a sequence of state-action pairs, $\xi=\{\left(\mathbf{s}_0,\mathbf{a}_0\right),\left(\mathbf{s}_1,\mathbf{a}_1\right),\dots,\left(\mathbf{s}_T,\mathbf{a}_T\right)\}$. The objective of reinforcement learning (RL) is to maximize the expected cumulative reward over trajectories sampled from the policy $\xi\sim p\left(\xi|\pi_\theta\right)$:
\begin{equation}
\label{eq:J_RL}
    \mathcal{J}_\text{RL}\left(\pi_\theta\right) = \mathbb{E}_{\xi}\left[\sum_{t=0}^T r\left(\mathbf{s}_t,\mathbf{a}_t\right)\right].
\end{equation}
There are various methods to train the policy in RL, with one popular approach being \textit{policy gradient} algorithms~\cite{silver2014deterministic}, such as REINFORCE~\cite{williams1992simple}. These methods update the policy parameters $\theta$ in the direction of the objective gradient:
\begin{equation}
\label{eq:J_RL_Grad}
\nabla_\theta\mathcal{J}_\text{RL}\left(\pi_\theta\right)=\mathbb{E}_\xi\left[\sum_{t=0}^T \nabla_\theta\log\pi_\theta\left(\mathbf{a}_t|\mathbf{s}_t\right)Q^{\pi_\theta}\left(\mathbf{s}_t,\mathbf{a}_t\right)\right],
\end{equation}
where $Q^{\pi_\theta}$ is the state-action value function (also known as the $Q$-function) estimator~\cite{sutton1999policy}.

\subsection{Diffusion Model and Diffusion Policy}\label{sec:dm_and_dp}
Denoising diffusion probabilistic models (DDPM)~\cite{ho2020denoising, sohl2015deep} are used to model the distribution of a dataset of samples, $\mathbf{x}^0$, conditioned on some context $\mathbf{c}$, represented as $\mathbf{x}^0 \sim p\left(\mathbf{x}^0|\mathbf{c}\right)$, where $\mathbf{x}^0 \in \mathbb{R}^n$. This conditional distribution is learned by modeling the reverse denoising process of a Markovian \textit{forward process} $q\left(\mathbf{x}^k|\mathbf{x}^{k-1}\right)$, which progressively adds Gaussian noise to the data samples over time.

The \textit{reverse process} $p\left(\mathbf{x}^{k-1}|\mathbf{x}^k, \mathbf{c}\right)$ is designed to recover the original, noise-free sample $\mathbf{x}^0$ from an initial Gaussian noise $\mathbf{x}^K \sim \mathcal{N}\left(0, \mathbf{I}\right)$ through $K$ iterations of denoising. This process generates a series of intermediate samples with progressively less noise, denoted as $\{\mathbf{x}^K, \mathbf{x}^{K-1}, \dots, \mathbf{x}^0\}$. Specifically, the reverse process is defined as
\begin{equation}
\label{eq:DM_reverse}
    p\left(\mathbf{x}^{k-1}|\mathbf{x}^k, \mathbf{c}\right) = \mathcal{N}\left(\mathbf{x}^{k-1}; \mu_\theta\left(\mathbf{x}^k, \mathbf{c}, k\right), \sigma_k^2 \mathbf{I}\right),
\end{equation}
where $\mu_\theta$ is a neural network parameterized by $\theta$ that predicts the added noise at each iteration, and $\sigma_k$ represents the step-dependent variance governed by a variance schedule.

The noise predictor $\mu_\theta$ is trained with the following objective:
\begin{equation}
    \mathcal{L}_\text{DM}\left(\theta\right) = \mathbb{E}_{\left(\mathbf{x}^0,\mathbf{c},\mathbf{x}^k,k\right)}\|\bar{\mu}\left(\mathbf{x}^0,k\right)-\mu_\theta\left(\mathbf{x}^k,\mathbf{c},k\right)\|^2,
\end{equation}
where $\bar{\mu}$ is the posterior mean of the forward process.

The diffusion policy (DP) models visuomotor robot policies using DDPMs, incorporating two key modifications: 1) The predicted data sample represents an \textit{action sequence} $\mathbf{A}_t$ of length $T_a$, defined as the action execution horizon; 2) The latest $T_s$ steps of \textit{state sequence} $\mathbf{S}_t$ at the time step $t$ is used as the conditional context for the denoising process.

Given $\mathbf{S}_t$, the conditional distribution of $\mathbf{A}_t$ is recovered through $K$ steps of reverse process, using a modified version of Eq.~\ref{eq:DM_reverse}:
\begin{equation}
    p\left(\mathbf{A}^{k-1}_t|\mathbf{A}^k_t, \mathbf{S}_t\right) = \mathcal{N}\left(\mathbf{A}^{k-1}_t; \mu_\theta\left(\mathbf{A}^k_t, \mathbf{S}_t, k\right), \sigma_k^2 \mathbf{I}\right).
\end{equation}
The noise predictor $\mu_\theta$ is trained with a modified $\mathcal{L}_\text{DM}$, defined as:
\begin{equation}
     \mathcal{L}_\text{DP}\left(\theta\right) = \mathbb{E}_{\left(\mathbf{A}_t,\mathbf{S},\mathbf{A}^k_t,k\right)}\|\bar{\mu}\left(\mathbf{A}_t,k\right)-\mu_\theta\left(\mathbf{A}^k_t,\mathbf{S},k\right)\|^2,
\end{equation}
where $\mathbf{A}^0_t$ is shorthand for $\mathbf{A}_t$, representing the final action sequence for execution.
\section{Fine-tuning Diffusion Policy with Human Preference}\label{sec:finetune}
In this section, we describe our approach for online RL-based fine-tuning of diffusion policy to align with human preference. First, a reward function representing human preference is obtained through preference-based reward learning (Sec.~\ref{sec:reward_learning}). Then, the reward model is used to fine-tune the diffusion policy (Sec.~\ref{sec:RL_finetuning}) using RL. We also incorporate KL regularization to stabilize fine-tuning, preventing over-fitting to human preferences while preserving the model’s ability to solve the original task (Sec.~\ref{sec:KL_regularization}).

\subsection{Preference-based Reward Learning}\label{sec:reward_learning}
The reward function estimator $\widehat{r}_\psi$ can be seen as encapsulating human judgments about various robot behaviors. This follows a standard framework where a reward function is trained to align with human preference labels~\cite{christiano2017deep,lee2021pebble,wang2024rl}. In this setup, a segment is defined as a sequence of states $\sigma = \{\mathbf{s}_1, \mathbf{s}_2, \dots, \mathbf{s}_H\}$, where $H \geq 1$. In our case, we consider $H = 1$, meaning each segment consists of a single state. For a pair of segments $\left(\sigma^0, \sigma^1\right)$, a human annotator provides a feedback label $y \in \{-1, 0, 1\}$, indicating which segment is preferred, where $0$ means the segment $\sigma^0$ is preferred, $1$ means the segment $\sigma^1$ is preferred, and $-1$ means both segments are equally preferable.

Using the Bradley-Terry model~\cite{bradley1952rank}, which assumes the probability of preferring one segment over another is exponentially dependent on the sum of an underlying reward function over the segment, the preference probability for a pair of segments, given the parameterized reward estimator $\widehat{r}_\psi$, can be expressed as
\begin{equation}
\label{eq:pref_prob}
    p_\psi[\sigma^1\succ \sigma^0] = \frac{\exp\left(\sum_{h=1}^H\widehat{r}_\psi\left(\sigma^1\right)\right)}{\sum_{i\in\{0,1\}}\exp\left(\sum_{h=1}^H\widehat{r}_\psi\left(\sigma^i\right)\right)},
\end{equation}
where $\sigma^i \succ \sigma^j$ denotes segment $\sigma^i$ being preferred over $\sigma^j$. In our case, Eq.~\ref{eq:pref_prob} simplifies to
\begin{equation}
    p_\psi[\mathbf{s}^1\succ \mathbf{s}^0] = \frac{\exp\left(\widehat{r}_\psi\left(\mathbf{s}^1\right)\right)}{\sum_{i\in\{0,1\}}\exp\left(\widehat{r}_\psi\left(\mathbf{s}^i\right)\right)}.
\end{equation}

Given a dataset of preference labels $D=\{\left(\sigma_i^0,\sigma_i^1,y_i\right)\}$, the reward function $\widehat{r}_\psi$ can be optimized by minimizing the following loss:
\begin{equation}
\begin{aligned}
    \mathcal{L}_\text{RWD}\left(\psi\right) =& -\mathbb{E}_{\left(\sigma^0,\sigma^1,y\right)}\Big[\mathds{1}\{y=\left(\sigma^0\succ\sigma^1\right)\}\log p_\psi[\sigma^0\succ \sigma^1]\\
    & +\mathds{1}\{y=\left(\sigma^1\succ\sigma^0\right)\}\log p_\psi[\sigma^1\succ \sigma^0]\Big],
\end{aligned}
\end{equation}
where $\mathds{1}\{\cdot\}$ equals to $1$ if the statement inside is true, and equals to $0$ otherwise.

In the setting of diffusion policy, at any time step $ t $, we generate an action sequence $ \mathbf{A}_t = \{\mathbf{a}_{t}, \mathbf{a}_{t+1}, \dots, \mathbf{a}_{t+T_a-1}\} $ using the diffusion policy conditioned on the state sequence $ \mathbf{S}_t = \{\mathbf{s}_{t-T_s+1}, \mathbf{s}_{t-T_s+2}, \dots, \mathbf{s}_{t}\} $. Consequently, we can express the reward as a function of the state-action sequence pair $ \left(\mathbf{S}_t, \mathbf{A}_t\right) $ by
\begin{equation}
    r_\psi\left(\mathbf{S}_t,\mathbf{A}_t\right) = \sum_{j=1}^{T_a}\widehat{r}_\psi \left(\mathbf{s}_{t+j}\right),
\end{equation}
where each future state $ \mathbf{s}_{t+j} $ is obtained by rolling out the action sequence $ \mathbf{A}_t $ starting from the current state $ \mathbf{s}_t $. Specifically, the state $ \mathbf{s}_{t+j} $ is sampled according to the transition probability $ \mathbf{s}_{t+j} \sim p\left(\mathbf{s}_{t+j} \mid \mathbf{s}_{t+j-1}, \mathbf{a}_{t+j-1}\right) $.

\begin{figure*}[t]
    \centering
    \includegraphics[width=0.95\linewidth]{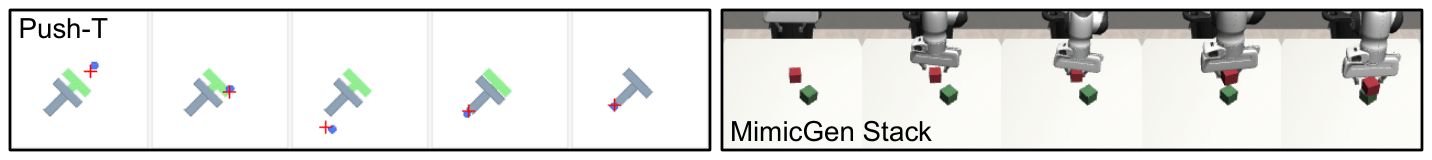}
    \caption{\textbf{Environments for Evaluation.} To evaluate the effectiveness of \textsc{FDPP}, We choose two long-horizon manipulation tasks including (left) a 2D pushing task \textsc{Push-T}~\cite{florence2022implicit,chi2023diffusion} and (right) a 3D pick-and-place task \textsc{Stacking} from \textsc{MimicGen}~\cite{mandlekar2023mimicgen}.}
    \label{fig:env}
\end{figure*}

\subsection{RL-based Fine-tuning}\label{sec:RL_finetuning}
Assume a pre-trained diffusion policy $p_\theta\left(\mathbf{A}_t^{0:K}|\mathbf{S}_t\right)$ is given. We can fine-tune this diffusion policy with the aforementioned reward function $r_\psi\left(\mathbf{S}_t,\mathbf{A}_t\right)$ by maximizing the denoising diffusion RL objective:
\begin{equation}
\label{eq:J_DDRL}
    \mathcal{J}_\text{DDRL}\left(\theta\right) = \mathbb{E}_{\left(\mathbf{S}_t,\mathbf{A}_t\right)}\left[r_\psi\left(\mathbf{S}_t,\mathbf{A}_t\right)\right],
\end{equation}
where $\mathbf{S}_t$ is obtained from roll-outs starting with an initial state sequence with padding $\mathbf{S}_0 = \{\mathbf{s}_0,\dots,\mathbf{s}_0\}$, where $\mathbf{s}_0\sim\rho_0$ follows the initial state distribution. The action sequence $\mathbf{A}_t$ is sampled through the pre-trained diffusion policy $\mathbf{A}_t\sim p_\theta\left(\mathbf{A}_t^{0:K}|\mathbf{S}_t\right)$. Note that we only keep the final action sequence $\mathbf{A}_t = \mathbf{A}_t^0$.

As introduced in~\cite{black2023training} and \cite{fan2024reinforcement}, we can represent the denoising process of DDPMs as a multi-step MDP, where the log-likelihood can be obtained via Monte-Carlo sampling. Specifically, we define the diffusion policy MDP $\mathcal{M}_\text{DP}$ as
\begin{align}
    \tilde{\mathbf{s}}_\tau &\triangleq \left(\mathbf{S}_t,k,\mathbf{A}_t^k\right),\\
    \tilde{\mathbf{a}}_\tau&\triangleq \mathbf{A}_t^{k-1},\\
    \pi\left(\tilde{\mathbf{a}}_\tau|\tilde{\mathbf{s}}_\tau\right) &\triangleq p_\theta\left(\mathbf{A}_t^{k-1}|\mathbf{A}_t^k,\mathbf{S}_t\right),\\
    \rho_0\left(\tilde{\mathbf{s}}_0\right)&\triangleq \left(p\left(\mathbf{S}_t\right),\delta_K,\mathcal{N}\left(\mathbf{0},\mathbf{I}\right)\right),\\
    p\left(\tilde{\mathbf{s}}_{\tau+1}|\tilde{\mathbf{s}}_\tau,\tilde{\mathbf{a}}_\tau\right) &\triangleq \left(\delta_{\mathbf{S}_t},\delta_{k-1},\delta_{\mathbf{A}_t^{k-1}}\right),\\
    r\left(\tilde{\mathbf{s}}_\tau,\tilde{\mathbf{a}}_\tau\right)&\triangleq \begin{cases}
        r_\psi\left(\mathbf{A}_t,\mathbf{S}_t\right)&\text{if}~\tau=0\\
        0&\text{otherwise}
    \end{cases}.
\end{align}
Here, $\delta_y$ represents the Dirac delta distribution, which has non-zero density only at $y$. The trajectories in the diffusion policy MDP $\mathcal{M}_\text{DP}$ consist of $K$ time steps, after which the state transition probability $p$ leads to a termination state. It is important to note that $\tau$ refers to the time step in $\mathcal{M}_\text{DP}$, $k$ refers to the denoising step in DDPM, and $t$ refers to the time step in the original environment where the diffusion policy is applied.

Since the cumulative reward of each trajectory $\tilde{\xi}$ in $\mathcal{M}_\text{DP}$ is equal to the final step reward $r_\psi\left(\mathbf{A}_t,\mathbf{S}_t\right)$, maximizing $\mathcal{J}_\text{DDRL}(\theta)$ in Eq.~\ref{eq:J_DDRL} is equivalent to maximizing $\mathcal{J}_\text{RL}(\pi)$ in Eq.~\ref{eq:J_RL}. Therefore, we can take the gradients with respect to the pre-trained diffusion policy parameters following Eq.~\ref{eq:J_RL_Grad}:
\begin{equation}
\begin{aligned}
    \nabla_\theta\mathcal{J}_\text{DDRL}&=\mathbb{E}_{\tilde{\xi}}\left[\sum_{\tau=0}^K \nabla_\theta\log\pi_\theta\left(\tilde{\mathbf{a}}_\tau|\tilde{\mathbf{s}}_\tau\right)Q^{\pi_\theta}\left(\tilde{\mathbf{s}}_\tau,\tilde{\mathbf{a}}_\tau\right)\right],\\
    &= \mathbb{E}\left[r_\psi\left(\mathbf{A}_t,\mathbf{S}_t\right)\sum_{k=1}^K\nabla_\theta\log p_\theta\left(\mathbf{A}_t^{k-1}|\mathbf{A}_t^k,\mathbf{S}_t\right)\right],
\end{aligned}
\end{equation}
where the expectation is taken over denoising trajectories generated by the current parameters $\theta$.

\subsection{KL Regularization}\label{sec:KL_regularization}
Fine-tuning a pre-trained diffusion policy solely using the preference-based reward model derived from human feedback risks over-fitting to the reward and forgetting the original task objective learned by the initial policy~\cite{fan2024reinforcement}. A common approach to mitigate this issue is to incorporate KL regularization~\cite{stiennon2020learning,fan2024reinforcement,ouyang2022training}. Specifically, we compute the KL divergence between the fine-tuned and pre-trained models for the final action sequence as a regularization term, \textit{i.e.}, $\mathcal{C} = \mathcal{D}_\text{KL}\left[p_\theta\left(\mathbf{A}_t|\mathbf{S}_t\right)\|p_\text{pre}\left(\mathbf{A}_t|\mathbf{S}_t\right)\right]$. Since obtaining a closed-form expression for $p_\theta\left(\mathbf{A}_t|\mathbf{S}_t\right)$ is challenging, we instead introduce an upper bound for this KL term into the objective function, following Lemma 4.2 in~\cite{fan2024reinforcement}:
\begin{equation}
\mathbb{E}_{\mathbf{S}_t}\Big[\mathcal{C}\Big]\leq \mathbb{E}_{\mathbf{S}_t}\left[\sum_{k=1}^K\mathbb{E}_{\mathbf{A}_t^k}\bigg(\overline{\mathcal{C}}\bigg)\right],
\end{equation}
where $\overline{\mathcal{C}} = \mathcal{D}_\text{KL}\Big[p_\theta\left(\mathbf{A}_t^{k-1}|\mathbf{A}_t^k,\mathbf{S}_t\right)\Big\|p_\text{pre}\left(\mathbf{A}_t^{k-1}|\mathbf{A}_t^k,\mathbf{S}_t\right)\Big]$. Therefore, we include the upper bound into Eq.~\ref{eq:J_DDRL} to get the new KL regularized objective function:
\begin{equation}
    \mathcal{J}_\text{DDRL}\left(\theta\right) = \mathbb{E}_{\mathbf{S}_t}\left[\mathbb{E}_{\mathbf{A}_t}\bigg(r_\psi\left(\mathbf{S}_t,\mathbf{A}_t\right)\bigg)+\alpha\sum_{k=1}^K\mathbb{E}_{\mathbf{A}_t^k}\bigg(\overline{\mathcal{C}}\bigg)\right],
\end{equation}
where $\alpha\geq 0$ is the weight of the KL term. Similarly, the new gradient is
\begin{equation}
\begin{split}
    &\nabla_\theta\mathcal{J}_\text{DDRL}= \mathbb{E}\Bigg[r_\psi\left(\mathbf{A}_t,\mathbf{S}_t\right)\sum_{k=1}^K\nabla_\theta\log p_\theta\left(\mathbf{A}_t^{k-1}|\mathbf{A}_t^k,\mathbf{S}_t\right)\\
    &+\alpha\sum_{k=1}^K\nabla_\theta\mathcal{D}_\text{KL}\Big[p_\theta\left(\mathbf{A}_t^{k-1}|\mathbf{A}_t^k,\mathbf{S}_t\right)\Big\|p_\text{pre}\left(\mathbf{A}_t^{k-1}|\mathbf{A}_t^k,\mathbf{S}_t\right)\Big]\Bigg].
\end{split}
\end{equation}

\subsection{Implementation Details}\label{sec:implementation_details}
We apply Proximal Policy Optimization (PPO)~\cite{schulman2017proximal} for the policy gradient update in the RL-based fine-tuning, which is more robust than the vanilla policy gradient methods. Diffusion policies are typically trained with stochastic sampler (\textit{e.g.}, DDPMs) with large sampling steps $K$. During fine-tuning, we use the Denoising Diffusion Implicit Model (DDIM)~\cite{song2020denoising} to reduce the number of sampling steps. One can change the deterministic level of DDIM through $\eta$, which controls the amount of noise injected into the sampling process, with $0$ being fully deterministic and $1$ being equivalent to the DDPM sampler. In practice, we set $\eta=1$ with $K^\text{DDIM}=50$ to improve the fine-tuning efficiency.

\section{Experimental Evaluation}\label{sec:experiments}

The purpose of our experiments is to evaluate the effectiveness of \textsc{\textbf{FDPP}} for fine-tuning diffusion policies to align with a variety of user-specified objectives. We focus on the following questions:
\begin{enumerate}
    \item Can \textsc{\textbf{FDPP}} align the pre-trained diffusion policy with human preference? (Sec.~\ref{sec:exp_q1})
    \item Does fine-tuning affect the performance of the original policy? (Sec.~\ref{sec:exp_q2})
    \item How does KL regularization help preserve original task objective during fine-tuning? (Sec.~\ref{sec:exp_q3})
\end{enumerate}
\begin{figure}[t]
    \centering
    \begin{minipage}[b]{0.49\linewidth}
        \centering
        \includegraphics[width=1.0\linewidth]{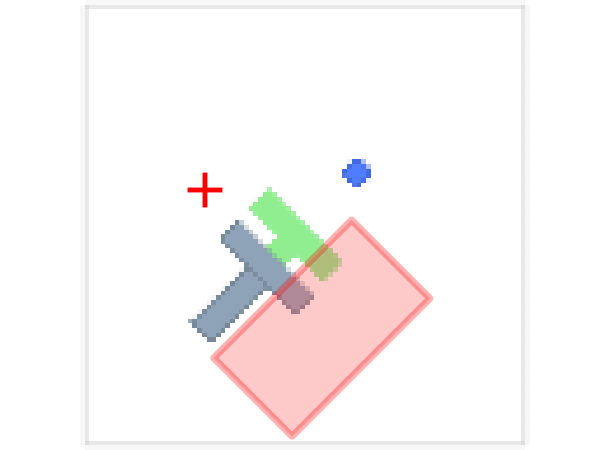} 
        \caption*{(a) Additional Constraints}
    \end{minipage}
    \hfill
    \begin{minipage}[b]{0.49\linewidth}
        \centering
        \includegraphics[width=1.0\linewidth]{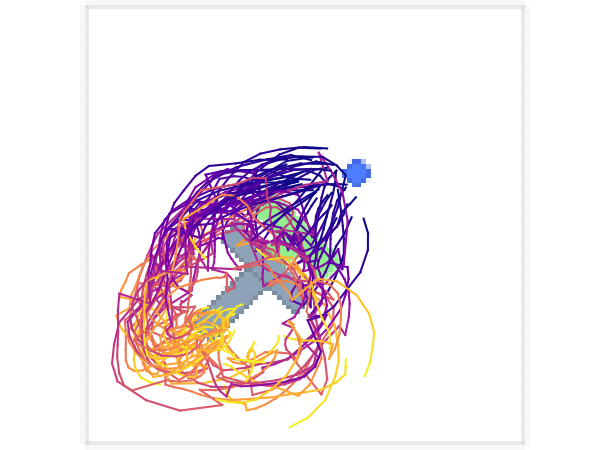} 
        \caption*{(b) Roll-out Trajectories}
    \end{minipage}

    \begin{minipage}[b]{0.49\linewidth}
        \centering
        \includegraphics[width=1.0\linewidth]{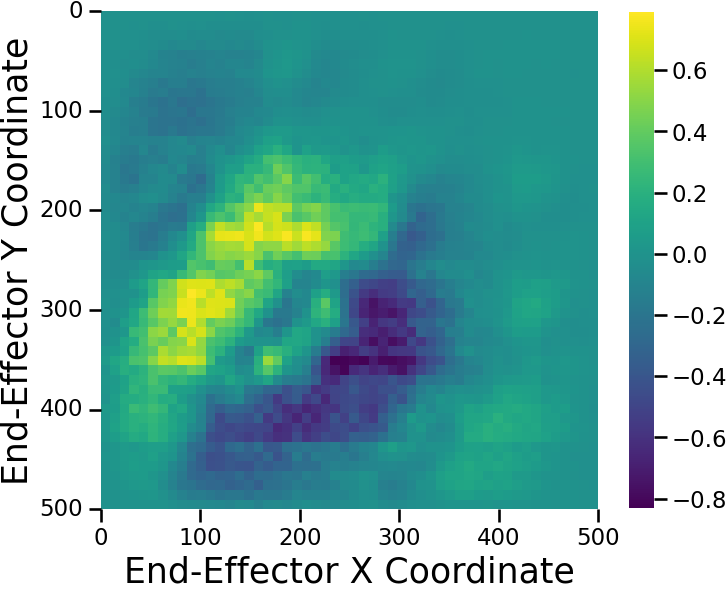} 
        \caption*{(c) Reward Distribution}
    \end{minipage}
    \hfill
    \begin{minipage}[b]{0.49\linewidth}
        \centering
        \includegraphics[width=1.0\linewidth]{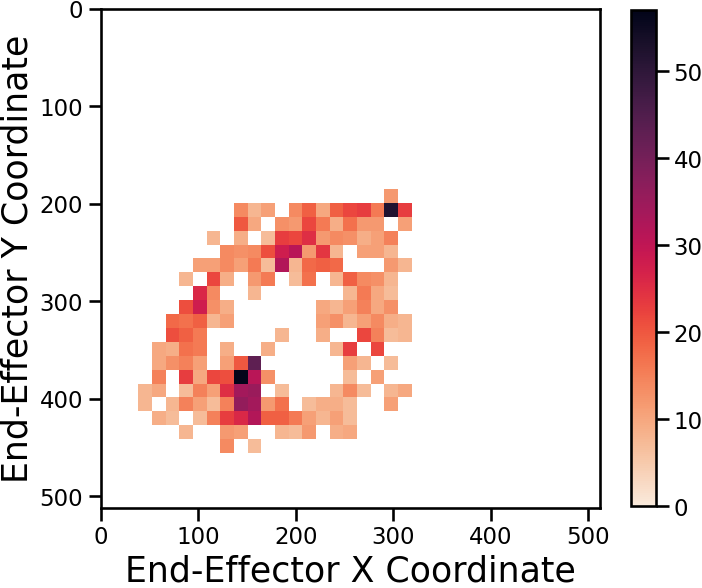} 
        \caption*{(d) Sample Frequency}
    \end{minipage}
    \caption{\textbf{Preference-based Reward Model.} Incorporating the additional constraints on the end-effector's location (Top-Left), where the red box represents the undesirable region, we perform roll-outs of the diffusion policy to gather trajectory samples (Top-Right). A reward model is then trained as described in Sec.~\ref{sec:reward_learning}, assigning varying values to different end-effector locations (Bottom-Left). Locations outside the sample distribution result in reward values remaining near zero (Bottom-Right).}
    \label{fig:reward}
\end{figure}

\begin{table}[t]
    \centering
    \scriptsize
    \caption{\textbf{Preference Alignment.} \textsc{\textbf{FDPP}} successfully adjusts the pre-trained diffusion policy to match human preferences for the desired feature. We present both the average feature over the entire trajectory and the feature at the terminal state.}
    \begin{tabular}{ccccc}
        \toprule
         & \multicolumn{2}{c}{Average}  & \multicolumn{2}{c}{Terminal State}  \\ \cmidrule(lr){2-3} \cmidrule(lr){4-5}
         & \textit{Pre-trained} & \textit{Fine-tuned} & \textit{Pre-trained}  & \textit{Fine-tuned} \\
        \midrule
        \color{myorange}{\textsc{\textbf{Push-T}}} [\%] & 0 & \textbf{100} & -- & -- \\
        \color{myblue}{\textsc{\textbf{Stack-Dist}}} [cm] & 6.28 & \textbf{3.17} & 3.91 & \textbf{0.94}\\
        \color{mypurple}{\textsc{\textbf{Stack-Align}}} [$^\circ$] & 36.73 & \textbf{26.16} & 32.97 & \textbf{21.20}\\
        \bottomrule
    \end{tabular}
    \label{tab:preference_alignment}
\end{table}

\subsection{Setup}
We choose two long-horizon manipulation tasks to evaluate \textsc{\textbf{FDPP}} as shown in Fig.~\ref{fig:env}. For each environment, we train a CNN-based diffusion policy following~\cite{chi2023diffusion} with pre-collected human demonstrations as the pre-trained policy.

\subsubsection{\color{myorange}{\textsc{\textbf{Push-T}}}} This environment is adapted from IBC~\cite{florence2022implicit}, where the task is to push a T-shaped block (gray) to a designated target location (green) using a circular end-effector (blue). The action space consists of the 2D planar position of the end-effector, while the observation space is $96\times96$ RGB image captured from a top-down view of the workspace.

We introduce an additional constraint for fine-tuning, ensuring that the end-effector does not enter an undesirable area located in the bottom-right of the workspace, depicted as the red box in Fig.~\ref{fig:reward}(a).

\subsubsection{\textsc{Stack}} This environment is adapted from \textsc{MimicGen}~\cite{mandlekar2023mimicgen}, which is a large-scale robotic manipulation benchmark for imitation learning and offline RL. We choose the block stacking task where the robot is required to pick up the red block and stack it onto the green block. The action space consists of the target joint angles of the 7-DoF Panda robot, while the observation space is the low dimensional state representation of the environment.

We introduce two additional preferences: 1) {\color{myblue}{\textsc{\textbf{Stack-Dist}}}}, which prefers minimizing the horizontal displacement between the centers of the red and green blocks; and 2) {\color{mypurple}{\textsc{\textbf{Stack-Align}}}}, which encourages reducing the misalignment angle between the red and green blocks.

For each environment, we obtain a diffusion policy following~\cite{chi2023diffusion} as the pre-trained policy.

\begin{figure}[t]
    \centering
    \begin{minipage}[b]{0.32\linewidth}
        \centering
        \includegraphics[width=1.0\linewidth]{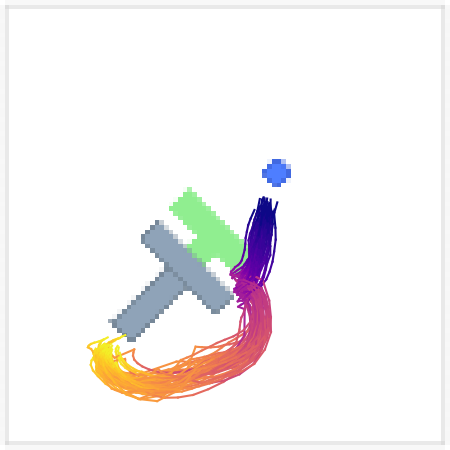} 
        \includegraphics[width=1.0\linewidth]{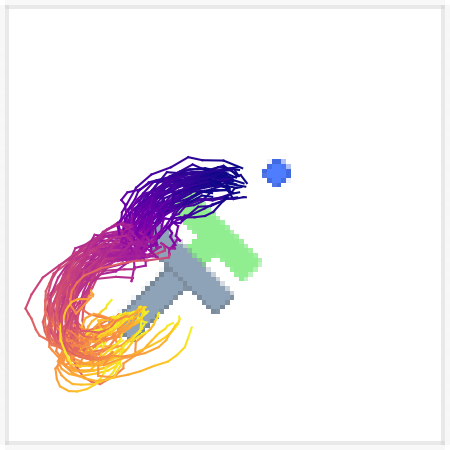} 
        \caption*{\color{myorange}{\textsc{\textbf{Push-T}}}}
    \end{minipage}
    \hfill
    \begin{minipage}[b]{0.32\linewidth}
        \centering
        \includegraphics[width=1.0\linewidth]{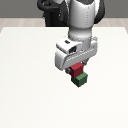} 
        \includegraphics[width=1.0\linewidth]{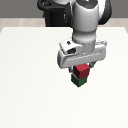} 
        \caption*{\color{myblue}{\textsc{\textbf{Stack-Dist}}}}
    \end{minipage}
    \hfill
    \begin{minipage}[b]{0.32\linewidth}
        \centering
        \includegraphics[width=1.0\linewidth]{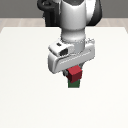} 
        \includegraphics[width=1.0\linewidth]{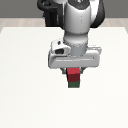} 
        \caption*{\color{mypurple}{\textsc{\textbf{Stack-Align}}}}
    \end{minipage}

    \caption{\textbf{Change of Behaviors.} \textsc{\textbf{FDPP}} can effectively shift the behavior distribution of the pre-trained diffusion policy (Top) to align with additional constraints or human preferences (Bottom).}
    \label{fig:pretrain_vs_finetune}
\end{figure}

\begin{table}[t]
    \centering
    \scriptsize
    \caption{\textbf{Effect of Fine-tuning on Policy Performance.} The impact of \textsc{\textbf{FDPP}} on the performance of the fine-tuned policy varies depending on the specific preferences being incorporated. The average roll-out length of \textsc{Stack} is reported as box-lifted-length / total-trajectory-length.}
    \begin{tabular}{ccccc}
        \toprule
         & \multicolumn{2}{c}{Success Rate}  & \multicolumn{2}{c}{Average Roll-out Length}  \\ \cmidrule(lr){2-3} \cmidrule(lr){4-5}
         & \textit{Pre-trained} & \textit{Fine-tuned} & \textit{Pre-trained}  & \textit{Fine-tuned} \\
        \midrule
        \color{myorange}{\textsc{\textbf{Push-T}}} &  \textbf{98\%} & 90\% & \textbf{58.20} & 120.20 \\
        \color{myblue}{\textsc{\textbf{Stack-Dist}}}  & 88\% & \textbf{92\%} & \textbf{36.20/121.80} & 37.32/131.28\\
        \color{mypurple}{\textsc{\textbf{Stack-Align}}}  & 88\% & \textbf{96\%} & 36.20/121.80 & \textbf{45.52/118.16}\\
        \bottomrule
    \end{tabular}
    \label{tab:performance}
\end{table}

\begin{figure*}[t]
    \centering
    \begin{minipage}[b]{0.119\linewidth}
        \centering
        \includegraphics[width=1.0\linewidth]{figures/pusht_pretrain.png} 
        \caption*{Pre-trained}
    \end{minipage}
    \begin{minipage}[b]{0.119\linewidth}
        \centering
        \includegraphics[width=1.0\linewidth]{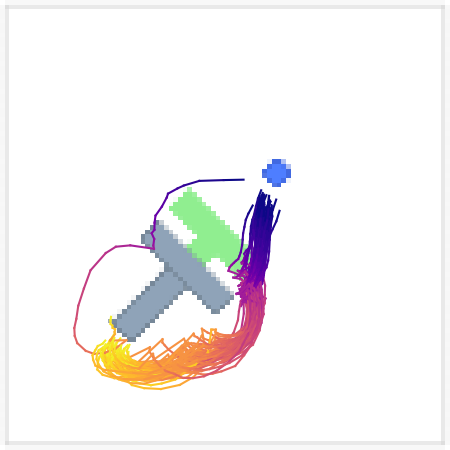} 
        \caption*{$\alpha=0.5$}
    \end{minipage}
    \begin{minipage}[b]{0.119\linewidth}
        \centering
        \includegraphics[width=1.0\linewidth]{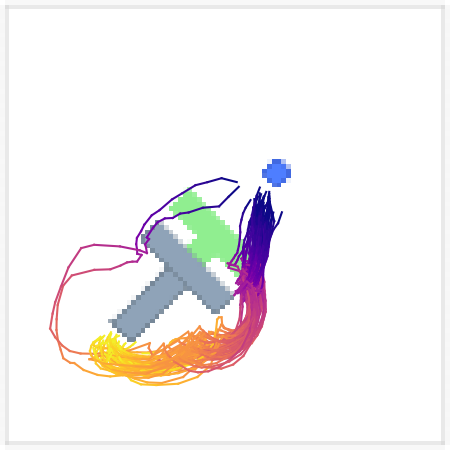} 
        \caption*{$\alpha=0.2$}
    \end{minipage}
    \begin{minipage}[b]{0.119\linewidth}
        \centering
        \includegraphics[width=1.0\linewidth]{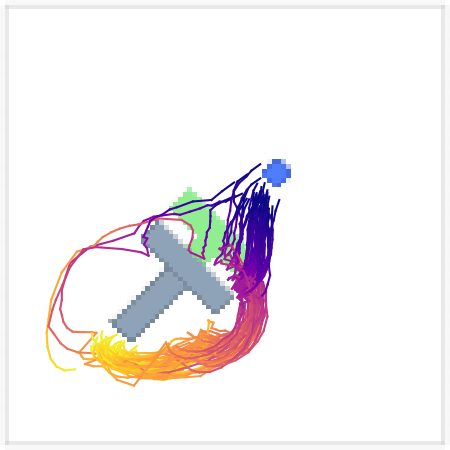} 
        \caption*{$\alpha=0.1$}
    \end{minipage}
    \begin{minipage}[b]{0.119\linewidth}
        \centering
        \includegraphics[width=1.0\linewidth]{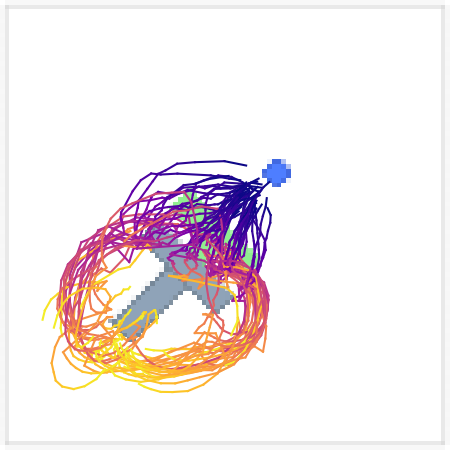} 
        \caption*{$\alpha=0.05$}
    \end{minipage}
    \begin{minipage}[b]{0.119\linewidth}
        \centering
        \includegraphics[width=1.0\linewidth]{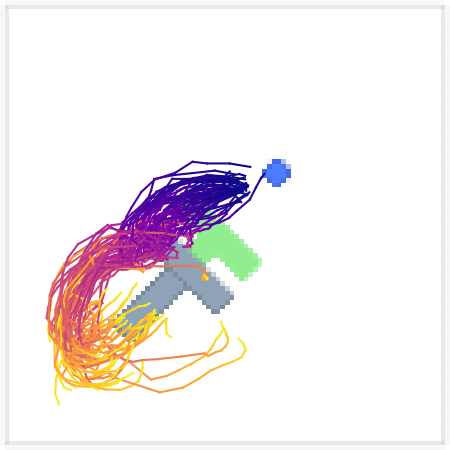} 
        \caption*{$\alpha=0.02$}
    \end{minipage}
    \begin{minipage}[b]{0.119\linewidth}
        \centering
        \includegraphics[width=1.0\linewidth]{figures/pusht_finetune_0.01.png} 
        \caption*{$\alpha=0.01$}
    \end{minipage}
    \begin{minipage}[b]{0.119\linewidth}
        \centering
        \includegraphics[width=1.0\linewidth]{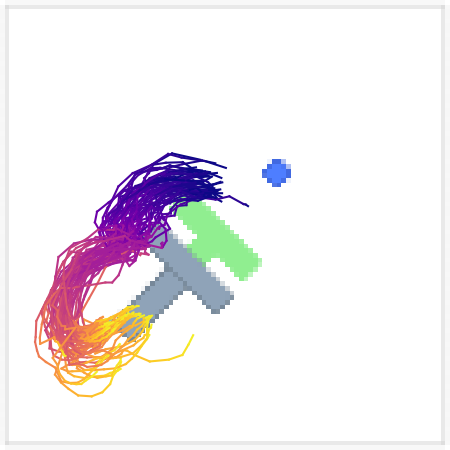} 
        \caption*{$\alpha=0.0$}
    \end{minipage}

    \caption{\textbf{Effect of KL Regularization on Policy Behavior.} A large KL regularization weight results in minimal deviation from the pre-trained policy. Reducing the KL weight allows the fine-tuned policy to better align with the reward model.}
    \label{fig:pusht_kl}
\end{figure*}

\begin{figure}[t]
    \centering
    \begin{minipage}[b]{0.49\linewidth}
        \centering
        \includegraphics[width=1.0\linewidth]{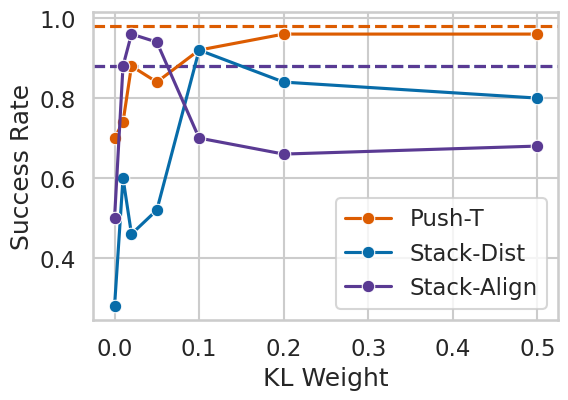} 
        \caption*{(a) Success Rate vs. $\alpha$}
    \end{minipage}
    \hfill
    \begin{minipage}[b]{0.49\linewidth}
        \centering
        \includegraphics[width=1.0\linewidth]{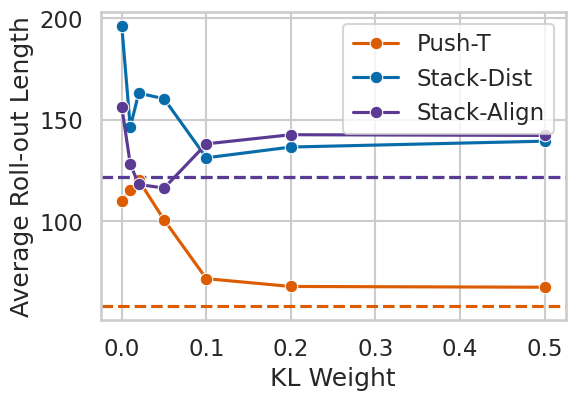} 
        \caption*{(b) Average Roll-out Length vs. $\alpha$}
    \end{minipage}
    \caption{\textbf{Effect of KL Regularization on Policy Performance.} A small KL weight leads to a significant decline in policy performance. Increasing the KL weight helps the fine-tuned policy's performance become more similar to that of the pre-trained policy. Dashed line represents the pre-trained policy performance.}
    \label{fig:kl_lines}
\end{figure}

\subsection{Preference Alignment}\label{sec:exp_q1}

For fine-tuning, we first train the preference-based reward model as described in Sec.~\ref{sec:reward_learning}. Figure~\ref{fig:reward} presents the resulting reward model for {\color{myorange}{\textsc{\textbf{Push-T}}}}. The training samples are generated by rolling out the pre-trained policy in the simulation, as depicted in Fig.~\ref{fig:reward}(b) (showing the first 40 steps of each trajectory for the end-effector). A human annotator provides preference labels on randomly sampled state pairs (see Fig.~\ref{fig:reward}(d) for sample frequency at the end-effector location). Each reward model is trained with 1024 state pairs. The final reward model (Fig.~\ref{fig:reward}(c)) effectively penalizes the area enclosed by the red box (undesirable region) and encourages the end-effector to move to the other side. Locations outside the sample distribution result in reward values that remain near zero. The reward model training for \textsc{\textbf{Stack}} follows a similar approach. However, instead of labeling pairs of end-effector locations, we label pairs based on horizontal displacement for {\color{myblue}{\textsc{\textbf{Stack-Dist}}}} and pairs based on orientation angle differences for {\color{mypurple}{\textsc{\textbf{Stack-Align}}}}.

Figure~\ref{fig:pretrain_vs_finetune} shows the qualitative results of the fine-tuning process. Utilizing the preference-based reward, we can either completely alter the behavior distribution of the pre-trained policy in {\color{myorange}\textsc{\textbf{Push-T}}}, achieve a smaller block displacement in {\color{myblue}\textsc{\textbf{Stack-Dist}}}, or reduce block misalignment in {\color{mypurple}\textsc{\textbf{Stack-Align}}}.

Table~\ref{tab:preference_alignment} quantitatively measures the alignment between the fine-tuned policy and human preference. For {\color{myorange}\textsc{\textbf{Push-T}}}, it shows the percentage of trajectories entering the undesirable area. For {\color{myblue}\textsc{\textbf{Stack-Dist}}}/{\color{mypurple}\textsc{\textbf{Stack-Align}}}, it lists the average and final distances/orientation misalignment between blocks. We conclude that \textbf{FDPP} successfully adjusts the pre-trained diffusion policy to match human preferences for the desired feature.

\subsection{Fine-tuned Policy Performance}\label{sec:exp_q2}
Table~\ref{tab:performance} presents the success rate and average roll-out length for both pre-trained and fine-tuned policies across each environment. In {\color{myorange}\textsc{\textbf{Push-T}}}, success is defined as achieving 90\% coverage of the gray T-shaped block on the green target. In \textsc{\textbf{Stack}}, success is determined by the red block being successfully stacked on the green block. Each environment has a maximum of 200 steps.

We observe that the fine-tuned policy's performance varies based on the specific preferences incorporated. In {\color{myorange}\textsc{\textbf{Push-T}}}, performance slightly decreases, with a longer average roll-out length, as the end-effector must take an alternate route to avoid the undesirable area. However, in {\color{myblue}\textsc{\textbf{Stack-Dist}}} and {\color{mypurple}\textsc{\textbf{Stack-Align}}}, fine-tuning enhances the success rate and reduces the average roll-out length. This improvement occurs because the human preferences are well-aligned with the task objectives, allowing fine-tuning with these rewards to boost policy performance.

\subsection{KL Regularization}\label{sec:exp_q3}
To prevent over-fitting, we introduce the KL regularizer (see Sec.~\ref{sec:KL_regularization}). Figure~\ref{fig:pusht_kl} shows the effect of KL regularization on policy fine-tuning in {\color{myorange}\textsc{\textbf{Push-T}}}. A large KL regularization weight results in minimal deviation from the pre-trained policy, while reducing the KL weight allows the fine-tuned policy to better align with the reward model. However, as illustrated in Fig.~\ref{fig:kl_lines}, a small KL weight can cause a significant decrease in policy performance because reinforcement learning tends to over-fit to the reward function, forgetting the original task objective of the pre-trained policy. This highlights the importance of KL divergence, as the original objective is not included in the reward function used for fine-tuning. In {\color{myblue}\textsc{\textbf{Stack-Dist}}} and {\color{mypurple}\textsc{\textbf{Stack-Align}}}, the impact of the KL weight on policy performance is more complex. Therefore, choosing an appropriate KL weight is essential to balance preference alignment and policy performance.

\section{Discussion}\label{sec:discussion}
In this work, we introduce \textbf{FDPP} for fine-tuning pre-trained diffusion policies to align with human preferences. Through extensive evaluations on two robotic tasks and three sets of preferences, we demonstrate the effectiveness of our method in customizing policy behavior distribution. Additionally, we explore the impact of KL regularization and find that incorporating a properly weighted KL regularizer can fine-tune the policy while preserving the original task objective from the pre-trained model.

For future work, we aim to build on the current setup and address some limitations: 1) Utilize Vision Language Models (VLMs) to automatically generate preference labels based on a single text description of human preferences, reducing human effort; 2) Evaluate on more long-horizon robotic tasks through real-world experiments; 3) Implement automatic hyper-parameter tuning to simplify the fine-tuning process.


\bibliographystyle{IEEEtran}
\bibliography{main}

\end{document}